\documentclass{article}
\usepackage{M-PSI}
\usepackage{amsmath,amssymb,amsfonts}
\usepackage{algorithmic}
\usepackage{graphicx}
\usepackage{textcomp}
\usepackage{booktabs}
\usepackage{subcaption}  
\usepackage{xcolor}
\usepackage{url}
\usepackage{multirow}
\usepackage{enumitem}
\def\BibTeX{{\rm B\kern-.05em{\sc i\kern-.025em b}\kern-.08em
    T\kern-.1667em\lower.7ex\hbox{E}\kern-.125emX}}

\usepackage[switch]{lineno}

\author{Xi Chen\textsuperscript{1,2}, Julien Cumin\textsuperscript{1}, Fano Ramparany\textsuperscript{1}, \href{https://research.vaufreydaz.org/}{Dominique~Vaufreydaz\textsuperscript{2,~\small{\ExternalLink }}}\vspace{0.1cm}\\
{$^1$ Orange Innovation}\\
{$^2$ Univ. Grenoble Alpes, CNRS, Grenoble INP, LIG, 38000 Grenoble, France}\\ %
}

\title{Towards LLM-Powered Ambient Sensor Based Multi-Person Human Activity Recognition}

\usepackage[pdfencoding=auto, pdfusetitle]{hyperref}
\hypersetup{
  hidelinks,
  urlcolor=red,
  pdftitle={Generative Resident Separation and Multi-label Classification for Multi-person Activity Recognition},
  pdfauthor={Xi Chen, Julien Cumin, Fano Ramparany, Dominique Vaufreydaz},
  pdfkeywords={Human Activity Recognition, Large Language Models, Ambient Intelligence, Smart Homes.},
}

\author{Xi Chen\textsuperscript{1,2}, Julien Cumin\textsuperscript{1}, Fano Ramparany\textsuperscript{1}, \href{https://research.vaufreydaz.org/}{Dominique~Vaufreydaz\textsuperscript{2,~\small{\ExternalLink }}}\vspace{0.1cm}\\
{$^1$ Orange Innovation}\\
{$^2$ Univ. Grenoble Alpes, CNRS, Grenoble INP, LIG, 38000 Grenoble, France}\\ %
}

\removefirstpagelogo{}
\begin{document}

\begin{abstract}
Human Activity Recognition (HAR) is one of the central problems in fields such as healthcare, elderly care, and security at home. However, traditional HAR approaches face challenges including data scarcity, difficulties in model generalization, and the complexity of recognizing activities in multi-person scenarios. This paper proposes a system framework called LAHAR, based on large language models. Utilizing prompt engineering techniques, LAHAR addresses HAR in multi-person scenarios by enabling subject separation and action-level descriptions of events occurring in the environment. We validated our approach on the ARAS dataset, and the results demonstrate that LAHAR achieves comparable accuracy to the state-of-the-art method at higher resolutions and maintains robustness in multi-person scenarios.

\end{abstract}

\section{Introduction}
Over the past two decades, \textbf{Human Activity Recognition (HAR)} using sensor technology has garnered increasing attention due to its potential applications in healthcare, security surveillance, and smart home environments. While many existing HAR systems employ camera-based technologies \cite{arshad2022human, vrigkas2015review}, these methods often raise substantial privacy concerns, particularly in private settings. As a response, some researchers have explored wearable technologies \cite{zhang2022deep}, such as smartwatches and smartphones. However, the requirement for individuals to continuously carry these devices may compromise comfort and convenience. Consequently, ambient sensors have gained back prominence as a key solution in HAR, prized for their non-invasive while avoiding privacy concerns of cameras and microphones.

Ambient sensors can be strategically placed within environments to detect and log changes in the physical state, with each change defined as an event. Common types of ambient sensors include door, presence, temperature, energy consumption sensors, and so on. Given their limited sensing range, multiple sensors are typically installed throughout a space to achieve thorough sensing coverage. The interactions between humans and their surroundings, recorded by these sensors, can be furthermore analyzed to infer individual actions and activities. This technology is known as \textbf{Ambient Sensor-Based Human Activity Recognition (AHAR)}.

However, AHAR faces the following challenges:
\begin{itemize}
    \item \textbf{Data Collection}: Due to the high cost of setting up experimental environments and the sensitivity of personal daily living data, collecting ambient sensor datasets is often challenging.
    \item \textbf{Model Generalization}: Due to varying sensor setups and activity routines, models trained on specific datasets often struggle to transfer their capabilities to different environments or configurations.
    \item \textbf{Context Integration}: Contextual information like sensor locations, functions, time, environment, and user habits is crucial due to the simplicity of ambient sensor data. However, traditional deep learning methods often fail to efficiently encode this information, making HAR less precise and flexible.
    \item \textbf{Multi-Person Recognition}: In environments where multiple individuals are present, events triggered by different subjects will blend into a single event sequence, complicating the task of activity recognition.
    \item \textbf{Explainability}: Explainable HAR helps increase user trust, enhance user experience, and improve system personalization. However, the inference process of traditional deep learning models is not intuitively understandable and lacks explainability.
\end{itemize}

\begin{figure*}[t]
    \centering
    \includegraphics[width=1.0\linewidth]{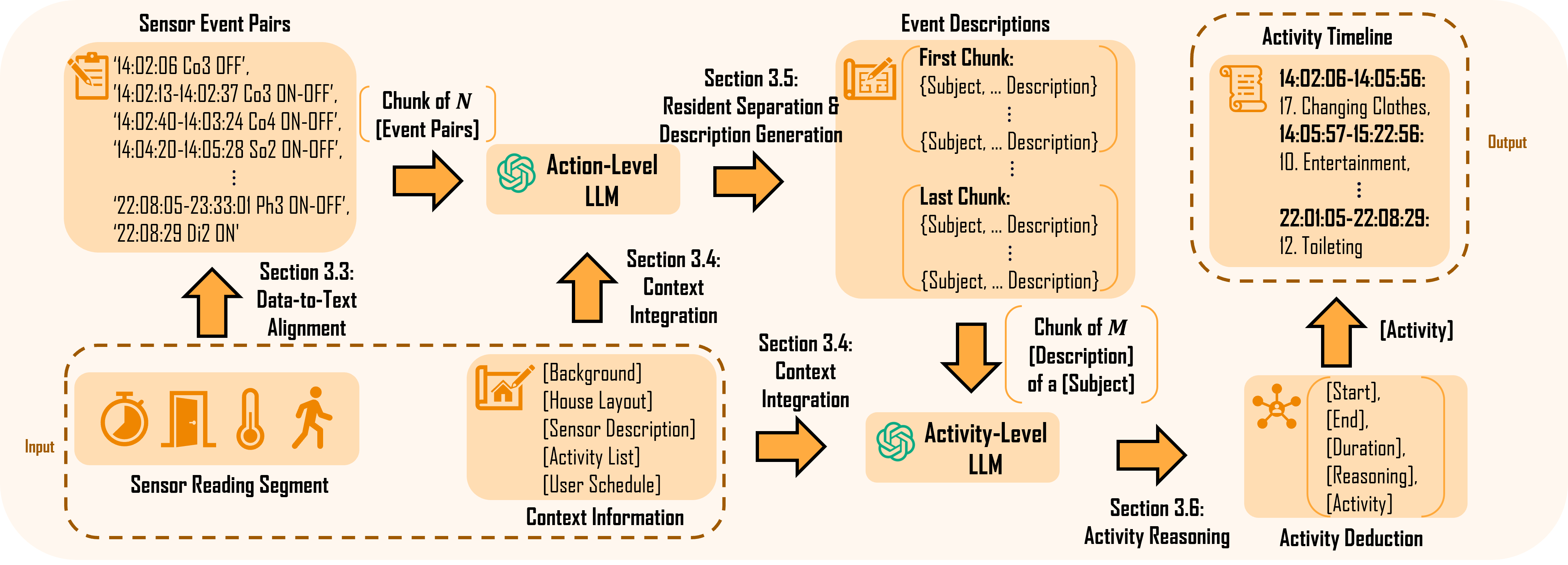}
    \caption{Workflow for our proposed LLM-based AHAR framework: LAHAR.}
    \label{fig:workflow}
\vspace{-1em}
\end{figure*}

In recent years, significant advancements have been made in \textbf{Large Language Models (LLMs)}, with models such as ChatGPT \cite{achiam2023gpt} and Llama \cite{touvron2023llama} exhibiting impressive contextual understanding and reasoning capabilities. This endows LLMs with the potential to address the aforementioned five challenges in AHAR: 1) LLMs' in-context learning capability \cite{brown2020language} reduces the need for training datasets; 2) by adapting relevant prompts, LLMs can swiftly adapt to novel environments or adjust to new sensor configurations; 3) leveraging the expressiveness of natural language, LLMs can integrate the different types of contextual information;
4) LLMs can connect related events using attention mechanisms, integrating common sense and reasoning to identify meaningful sensor event combinations. Furthermore, LLMs' generation capabilities allow them to generate regrouped coherent sequences. Therefore, the LLMs have the potential to separate mixed event sequences in multi-person scenarios;
5) LLMs possess the ability to explain their reasoning process, thereby enhancing the explainability of the inference.

The objective of this study is to leverage the advanced capabilities of LLMs to tackle the identified challenges in AHAR in an efficient and explainable way. To achieve this, we propose a two-stage framework, \textbf{LAHAR} (\textbf{L}LM-powered \textbf{AHAR}), designed to process multi-person sensor data from fine to coarse granularity, enabling few-shot learned recognition of activities. In the first stage, LAHAR is fed textualized sensor events at the level of seconds, assigns and describes each subject's actions at a fine granularity using natural language. Based on each subject's action descriptions collected from the first stage, LAHAR then performs reasoning in the second stage to predict a coarse timeline of activities spanning up to tens of hours.

The contributions of this study are listed as follows:
\begin{enumerate}
    \item We propose an LLM-based AHAR approach which can be applied in multi-person scenarios. To the best of our knowledge, this is the first approach employing LLMs for recognizing multi-person activities.
    \item We present a fine-to-coarse two-stage prompt engineering method that enables our system to continuously provide precise natural language descriptions of sensor data spanning over hours, and to further integrate these descriptions to infer daily living activities.
\end{enumerate}

\section{Related Work}
Recently, increasing attention has been given to modeling ambient sensor sequences using natural language models. Bouchabou et al. \cite{bouchabou2021fully, bouchabou2021using} first introduced the concept of language models into human activity recognition, treating each sensor event as a word (token), and used the word embedding method to learn the correlations between sensor events. Zhao et al. \cite{zhao2022feature} further encoded the sensor environmental location into the embedding vectors, demonstrating the capability of language models to integrate context information. Das et al. \cite{das2023explainable} elaborated on the importance of explainability in activity recognition and implemented a system capable of explaining activity recognition classifications using natural language. Takeda et al. \cite{takeda2023sensor} first used the large language model GPT2 \cite{radford2019language} for generative prediction of sensor event sequences, predicting future sensor events based on the labels of the ongoing activity and the sensor events that have already occurred. This work further strengthened the association of human activity recognition with language models and brought the large GPT model \cite{radford2018improving} into the scope of HAR.
\begin{figure*}[t]
    \centering
    \includegraphics[width=\linewidth]{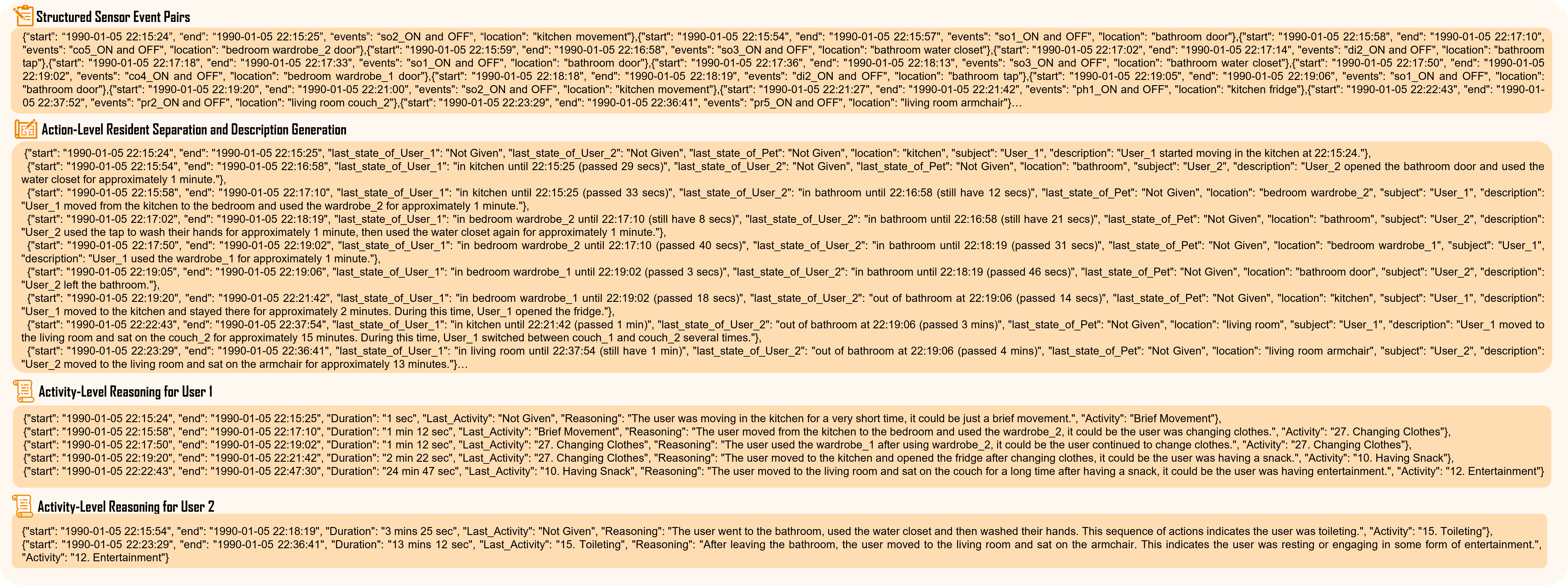}
\caption{Example of outputs generated by LAHAR at each stage}
\label{fig:qulitative_example}
\end{figure*}

With large language models demonstrating powerful in-context learning \cite{brown2020language} and reasoning \cite{wei2022chain} abilities, Gao et al. \cite{gao2024unsupervised} first used a large language model to perform unsupervised annotation on single-person activity samples in the ARAS dataset \cite{alemdar2013aras}, demonstrating the potential of large language models for unsupervised human activity recognition. In this work, Gao et al. used sensor reading data within a 5-minute sliding window as input data. They employed a Chain-of-Thought approach \cite{wei2022chain} to instruct the LLM to analyze the functions of the activated sensors. By integrating context information on room layout, time, and the duration of sensor activation, LLM was finally instructed to choose an activity as the recognition results from nine activities selected by the authors. Although the experimental results show comparable accuracy to supervised trained models, this work is limited to nine easily distinguishable activity categories in single-person scenarios, overlooking the recognition of other more challenging categories and failing to provide prompts for reproducibility.
Furthermore, using sensor readings in fixed time windows rather than sensor events as input data limits the model's ability to perceive the subject's behavior at a finer granularity.

Although language models are widely used in applications such as sensor representation, event sequence prediction, and activity explanation in single-user scenarios, these methods are often difficult to apply directly in multi-user scenarios. This is because only modeling the correlations of sensor events is insufficient to separate the activity information of different subjects in multi-person scenarios.
To separate sensor events from different subjects, Wang and Cook \cite{wang2020smrt, wang2021multi} first applied a skip-gram word embedding model to learn sensor correlations and then used a Gaussian Mixture Probability Hypothesis Density (GM-PHD) filter to cluster events into different tracks. Instead of using a probabilistic model, Chen et al. \cite{chen2024generative} employed a Sequence-to-Sequence model \cite{sutskever2014sequence}, using a machine-translation-like method, further applying language models to multi-person human activity recognition. This method first encoded the mixed event sequence of two subjects and then generatively decoded it into two single-person sequences separated by delimiters, thus achieving the final separation of multi-person event sequences. This research demonstrated the potential of generative methods for separating confounded information.

In this work, we leverage the powerful encoding capabilities of large language models, along with the separation abilities of generative methods, applying a generative LLM to multi-person activity recognition.

\section{Methodology}

Figure \ref{fig:workflow} illustrates the workflow of our proposed framework LAHAR. Given a time period $T$, the collection of all sensor readings within it is referred to as a \textit{sensor reading segment}. LAHAR includes three main steps of information processing: 1)~Process the sensor reading segment into a textual form of sensor event pairs (Section \ref{sec:Data-to-Text}); 2)~Integrate the context information (Section \ref{sec:Context_Integration})  into the sensor event pair sequence, separate subjects and generate individual action-level descriptions by an LLM (Section \ref{sec:Description_Generation}); 3)~Based on the action-level descriptions and the context information, an LLM is used to perform activity-level reasoning to predict the timeline of activities for each subject (Section \ref{sec:Activity_Reasoning}).

\subsection{Problem Formalization}
Given an environment $E=\{s_i\}_{1\leq i\leq n}$, where $s_i$ is a sensor installed in the environment characterized by its specific setting, we define a sensor event as $e_t=<t, s, c>$, where $t$ represents the time of the event, $s$ represents the sensor, and $c$ represents the change in sensor status. The sequence of events that occur within a time period $T=[t_s,t_e]$ is $S_T=(e_{t_1}, e_{t_2},...,e_{t_k})$ where $\forall i \in [1,k], t_i \in [t_s, t_e]$. Given an activity category set of $K$ activities $L_{A} = \{a_k\}_{1\leq k \leq K}$, the activities occurred during $T$ are defined as $A_T = \{a_{k}^{T_j}\}_{j\in \mathcal{J}}$, where $\mathcal{J} = \{j \in \mathbb{N} | T_j \subseteq T\}$. The objective of this research is to propose a model $M$ such that $A_T=M(S_T|E, L_A)$.

\subsection{Data-to-Text Alignment}\label{sec:Data-to-Text}
As LLMs accept text as input, LAHAR first involves data-to-text alignment. This process includes two steps: data preprocessing, and information structuring.

\paragraph{Data Preprocessing}
Unlike Gao et al. \cite{gao2024unsupervised}, who extract overall features from all sensor readings within a fixed time window, our method first preprocesses the sensor readings into sensor events. Specifically, when there is a change in the reading of any sensor, we denote the time of occurrence $t$, the changed sensor identifier $s$, and the change of sensor reading $c$ as a sensor event $e = <t, s, c>$. Since a sensor event often corresponds to an action by a subject, analyzing events allows our model to achieve fine-grained, action-level detail. Meanwhile, to reduce redundant information, when a sensor continuously changes at a high frequency between two states without any other sensor events occurring, we retain only the first and the last events.

\paragraph{Information Structuring}
To further enhance the information density and quality of the text input to the LLM, we perform information structuring on the sequence of sensor events. For adjacent activation $e_{ON} = <t_s, s, \text{ON}>$ and deactivation events $e_{OFF} = <t_e, s, \text{OFF}>$ of a sensor, we pair them into an event pair $p$, incorporating the sensor's location information. We then structure them into a JSON format as follows:
$p = \{``start": <t_s>, ``end": <t_e>, ``event": <s> \text{ON and OFF}, ``location": <l>\}$. In the end, all event pairs are provided to the LLM sorted in ascending order by start time. This structure of event pairs is designed to help the LLM identify residents' occupancy in multi-person scenarios. An example of final structured event pairs is illustrated in the first block of Figure \ref{fig:qulitative_example}.

\subsection{Context Integration} \label{sec:Context_Integration}
Traditional machine learning methods struggle with ambient sensor data due to limited information from sensor readings. However, contextual information like sensor location, type, function, user habits, and environment layout often provide more insight than the sensor data itself. Integrating this contextual information is crucial for understanding the correlations between sensors and for activity recognition.

Given that ambient sensors are usually installed in a relatively stable environment, this contextual information tends to remain constant. Therefore, our method proposes to provide contextual information to LLMs through language prompts, so that LLMs can harness their encoding capabilities to embed this information and align it with relevant sensor events. The contextual information used in this work is listed as follows:
\begin{itemize}
    \item \textbf{Background:} 
This introduces the role of the LLM, the number of residents, and the fact that ambient sensors are installed to identify activities.
    \item \textbf{House Layout:}
This provides the list of rooms contained in the environment, the furniture in each room, and the associated sensors.
    \item \textbf{Sensor Description:}
This explains the identifier, type, and location of each sensor in the environment.
    \item \textbf{Activity List:}
This offers a list of possible activities within the environment, along with certain behavior patterns or user habits related to these activities.
    \item \textbf{User Schedule:}
This emphasizes the intervals during which subjects perform certain activities, such as eating breakfast.
\end{itemize}

\subsection{Action-Level Resident Separation and Description Generation} \label{sec:Description_Generation}

This section depicts the design of an LLM-powered module that assigns the sequential event pairs from the Data-to-Text module (Section \ref{sec:Data-to-Text}) to different subjects, and then provides natural language descriptions with action-level granularity as illustrated in the second block of Figure \ref{fig:qulitative_example}. This process is primarily based on two assumptions: 1)~Related sensor events are more likely to be triggered by the same person; 2)~A person cannot simultaneously trigger two unrelated sensors. 

The first assumption enables the LLM to merge related sensor events, while the second assumption enables the LLM to separate events triggered by different subjects. The application of these two assumptions relies on two abilities: 1)~Determining the relevance of sensor events; 2)~Assessing the current state of the subject to determine their likelihood of triggering other events. 

The first ability can be enabled by the context introduced in Section \ref{sec:Context_Integration} and the common sense learned by the LLM during its training, indicating the use of the In-Context Learning \cite{brown2020language}. The second ability requires us to introduce a Chain of Thought \cite{wei2022chain} in the prompt to guide the reasoning of the LLM. Therefore, the prompt contains 4 basic components: 1)~\textit{Context}; 2)~\textit{Instructions}; 3)~\textit{Examples}; 4)~\textit{Input}, where \textit{Context} and \textit{Examples} follows the idea of In-Context Learning, and \textit{Instructions} describes the Chain of Thought.

\paragraph{Input}
Although the \textit{Input} section appears last in the prompt, we introduce it first for clarity. Given a period $T$, the sequence of events $S_T$ is formatted into a sequence of event pairs $P_T$ following Data-to-Text alignment. Since $P_T$ can be too long to ensure high-quality generation, we divide $P_T$ into chunks $C_i$, each containing $N$ event pairs, except for the last chunk, which contains the remaining pairs. We process each chunk sequentially in a loop, concatenating all responses at the end. To enable the LLM to infer the users' last state at the beginning of each new step, we include the final description of each subject from the previous chunk into the input of the subsequent step.
\paragraph{Context}
In this part of the prompt, \textit{Background}, \textit{House Layout}, and \textit{Sensor Description} introduced in Section \ref{sec:Context_Integration} are provided to the LLM as the context information. Formally, we have
$Context =  ([Background], [House Layout], [Sensor Description]).$

\paragraph{Instructions}
We instruct the LLM to sequentially perform the following steps: 
\begin{enumerate}
    \item Merge related sequential event pairs, and determine the overall start and end times;
    \item Summarize the previous action states of the users and determine whether the previous actions have ended;
    \item Recall the location of the current event pairs;
    \item Considering the previous states of users, designate a suitable user as the subject for the current event pair being processed;
    \item Describe the current event pair with natural language.
\end{enumerate}
Ultimately, the prompt ask the LLM to respond in a predefined JSON format, which implicitly formalizes the Chain of Thought while making the generated results easier to post-process and increasing the information density. The keys defined in the JSON format are: \{\textit{``start"}, \textit{``end"}, \textit{``last state of User 1"}, ..., \textit{``last state of User i"}, \textit{``location"}, \textit{``subject"}, and \textit{``description"}\}.

\paragraph{Examples}
To further activate the LLM's ability to use context and follow the chain of thought for reasoning, the prompt provides several examples to the LLMs.

\subsection{Activity-Level Reasoning} \label{sec:Activity_Reasoning}
The objective of this second module is to align descriptions of fine granular action of each subject to an activity timeline $A_T$ as shown in the last two blocks of Figure \ref{fig:qulitative_example}. For activities that are directly associated with sensors, LLMs can make use of common sense reasoning, such as associating sleeping with the pressure sensor of a bed. On the other hand, recognizing activities that are environment-specific and user-specific relies heavily on in-context learning. Consequently, the design of context and examples is crucial for this module.
Similar to the Description Generation module, the prompt contains 4 basic components: 1)~Context; 2)~Instructions; 3)~Examples; 4)~Input.

\paragraph{Input}
From the output of last module, we separate and reorganize the descriptions for each subject, retaining only 4 key-value pairs: ``start", ``end", ``location", and the ``description”. After implementing the separation, we input each subject's descriptions independently. Similarly to the previous module, we divide each subject's descriptions into chunks, with each containing $M$ descriptions.

\paragraph{Context}
In this part of the prompt, \textit{Sensor Description}, \textit{Activity List}, and \textit{User Schedule} introduced in Section \ref{sec:Context_Integration} are provided to the LLM as the context information. Formally, we have
$Context =  ([Sensor Description], [Activity List], [User Schedule]).$

\paragraph{Instructions}
We instruct the LLM to sequentially perform the following steps: 
\begin{enumerate}
    \item Analyse and summarise the descriptions that belong to the same activity, and determine the overall start and end times;
    \item Calculate the duration of the acitivty;
    \item Recall the last activity predicted;
    \item Considering the previous activities of the subject and the duration of current actions, reason the subject's current activity;
    \item Choose an activity with ID from the activity list.
\end{enumerate}
Ultimately, we instruct the LLM to respond in a predefined JSON format, in which the keys defined are: \{\textit{``start"}, \textit{``end"}, \textit{``Duration"}, \textit{``Last\_Activity"}, \textit{``Reasoning"}, and \textit{``Activity"}\}.

\paragraph{Examples}
For activities that cannot be directly detected by sensors, they are often described by multiple groups of sensor events and typically exhibit certain patterns. We filter out the corresponding descriptions for these activities, then provide correct reasoning results to explain why these descriptions correspond to the given activity.

\section{Experiments}
\subsection{Dataset}

To evaluate LAHAR, we require an ambient-based multi-person HAR dataset that provides sufficient contextual information for all sensors, enabling them to be described with language. To the best of our knowledge, the ARAS dataset \cite{alemdar2013aras} best meets this requirement. 
It is a publicly available dataset that includes two real houses (named House A and B), each equipped with 20 ambient sensors. Within each house, a maximum of two subjects can concurrently be observed engaging in any of 27 different daily activities.
Each house dataset contains 30 files, representing 30 days. Each daily file contains sensor reading data and multi-person activity annotations, both at the level of seconds, thus including 86400 annotated data instances.
\begin{table}[t]
  \centering
  \caption{Activity ID and labels of ARAS Datasets after regrouping.}
  \label{tab:activity_id}
  \resizebox{0.49\textwidth}{!}{  
  \begin{tabular}{clclclcl}
    \toprule
    ID & Activity & ID & Activity & ID & Activity & ID & Activity \\
    \midrule
    0 & Other & 1 & Preparing Breakfast & 2 & Having Breakfast &  3 & Preparing Lunch \\
    4 & Having Lunch & 5 & Preparing Dinner & 6 & Having Dinner & 7 & Washing Dishes \\
    8 & Having Snack & 9 & Sleeping & 10 & Entertainment & 11 & Having Shower \\
    12 & Toileting & 13 & Working & 14 & Shaving & 15 & Brushing Teeth \\
    16 & Talking on the Phone & 17 & Changing Clothes & & & & \\
    \bottomrule
  \end{tabular}
  }
\end{table}

\subsection{Experiment Settings}
\paragraph{Data Segmentation}
Although our method can reason coherently without prior data segmentation, we performed necessary segmentation. We noted that House A's data is daily independent, while House B's data spans 30 consecutive days. Thus, we concatenated House B's 30 days of data but treated each day in House A as an independent segment. For evaluation, we further divided the data into single-person and multi-person scenarios based on the "Leaving House" activity. Consequently, House A had 59 single-person and 61 multi-person segments, while House B had 10 single-person and 24 multi-person segments.

\paragraph{Error Preprocessing}

We assessed sensor error levels in the houses by examining the number of events that occurred when both residents were leaving the house. According to our observation, House A exhibited significant noise, especially from the hall motion sensor, kitchen motion sensor, and kitchen temperature sensor. To address this, we removed the hall motion sensor events and deactivated the kitchen motion and temperature sensors, except during kitchen activities.

\paragraph{Class Selection and Regrouping}

Due to similar activities in the ARAS dataset that sensors don't distinguish, we merged certain activities: \textit{Napping} and \textit{Sleeping} into \textit{Sleeping}, and \textit{Watching TV}, \textit{Reading books}, and \textit{Listening to music} into \textit{Entertainment}. In House A, \textit{Using Internet} and \textit{Studying} were merged into \textit{Working}; in House B, \textit{Using Internet} was merged into \textit{Entertainment}, and \textit{Studying} was renamed to \textit{Working}. We removed infrequent activities like \textit{Laundry}, \textit{Cleaning}, \textit{Having conversations}, and \textit{Having guests}, as recognizing these activities is beyond the capability of our method. This resulted in the list of activities shown in Table \ref{tab:activity_id}.

\begin{figure}
    \centering
    \begin{subfigure}[b]{0.49\linewidth}
    \includegraphics[width=\linewidth]{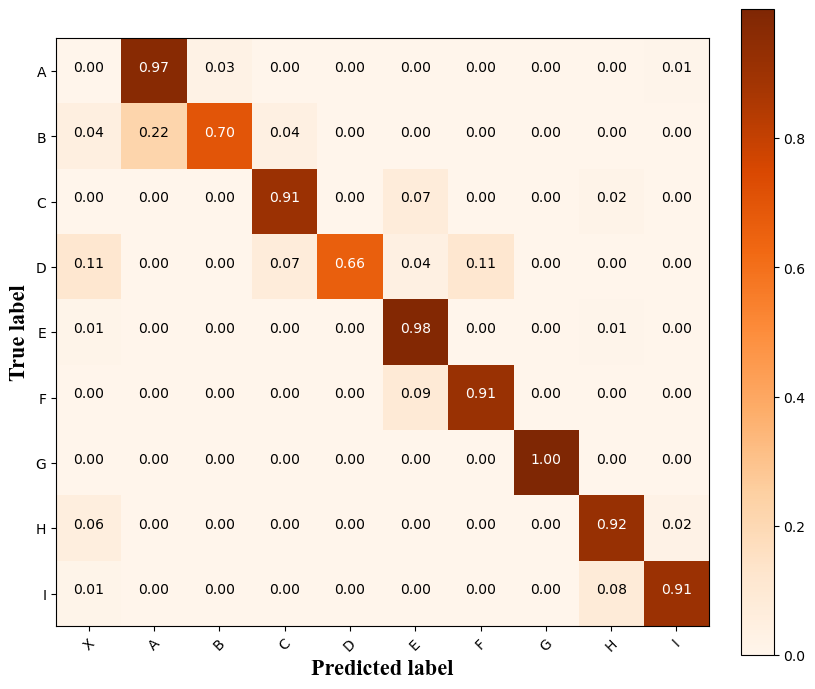}
    \caption{LAHAR: at the level of seconds}
    \label{fig:cm_ours}
    \end{subfigure}
\hfill
    \begin{subfigure}[b]{0.49\linewidth}
    \includegraphics[width=\linewidth]{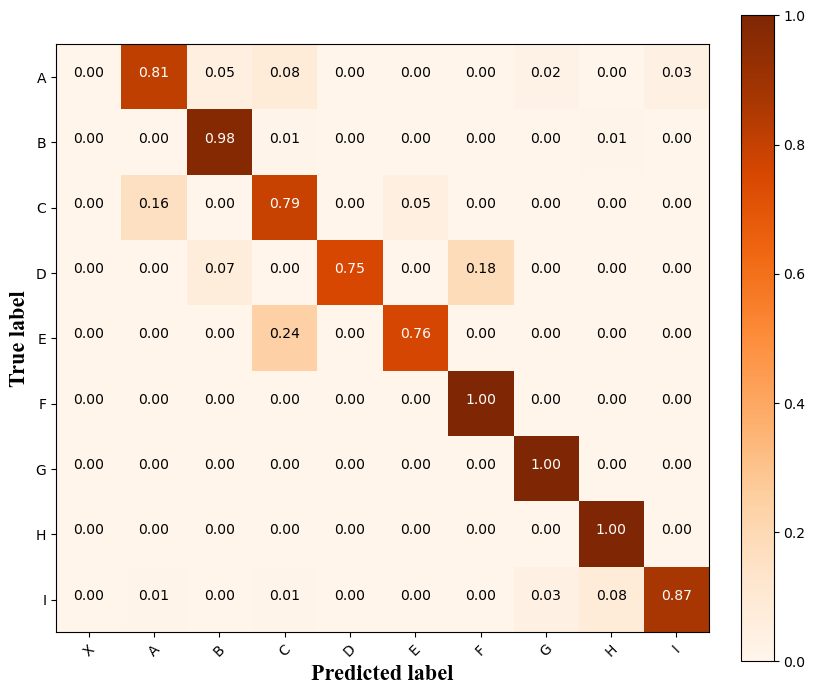}
    \caption{Gao et al. \cite{gao2024unsupervised}: at the level of 5 minutes}
    \label{fig:cm_Gao}
    \end{subfigure}
\caption{Comparison of confusion matrices between our method and the Gao et al. method \cite{gao2024unsupervised}. The categories are: X)~Unknown, A)~Preparing Breakfast, B)~Having Breakfast, C)~Preparing Lunch, D)~Having Lunch, E)~Preparing Dinner, F)~Having Dinner, G)~Sleeping, H)~Having Shower, I)~Toileting.}
\label{fig:cm_compare}
\end{figure}

\begin{table}[t]
\centering
\caption{Comparison of performance between our method and the state of the art}
\label{tab:comparison}
\begin{tabular}{l|cccc}
\hline
Method & Resolution & Precision & Recall & F1-score \\ \hline
Gao et al.\cite{gao2024unsupervised}     & 5 minutes  & 96.00 & 95.56 & 95.60                    \\ 
LAHAR     & Second   & 88.54 & 92.31 & 90.39               \\
\hline
\end{tabular}
\end{table}

\begin{figure}[t]
    \centering
    \begin{subfigure}[b]{0.49\linewidth}
    \includegraphics[width=\linewidth]{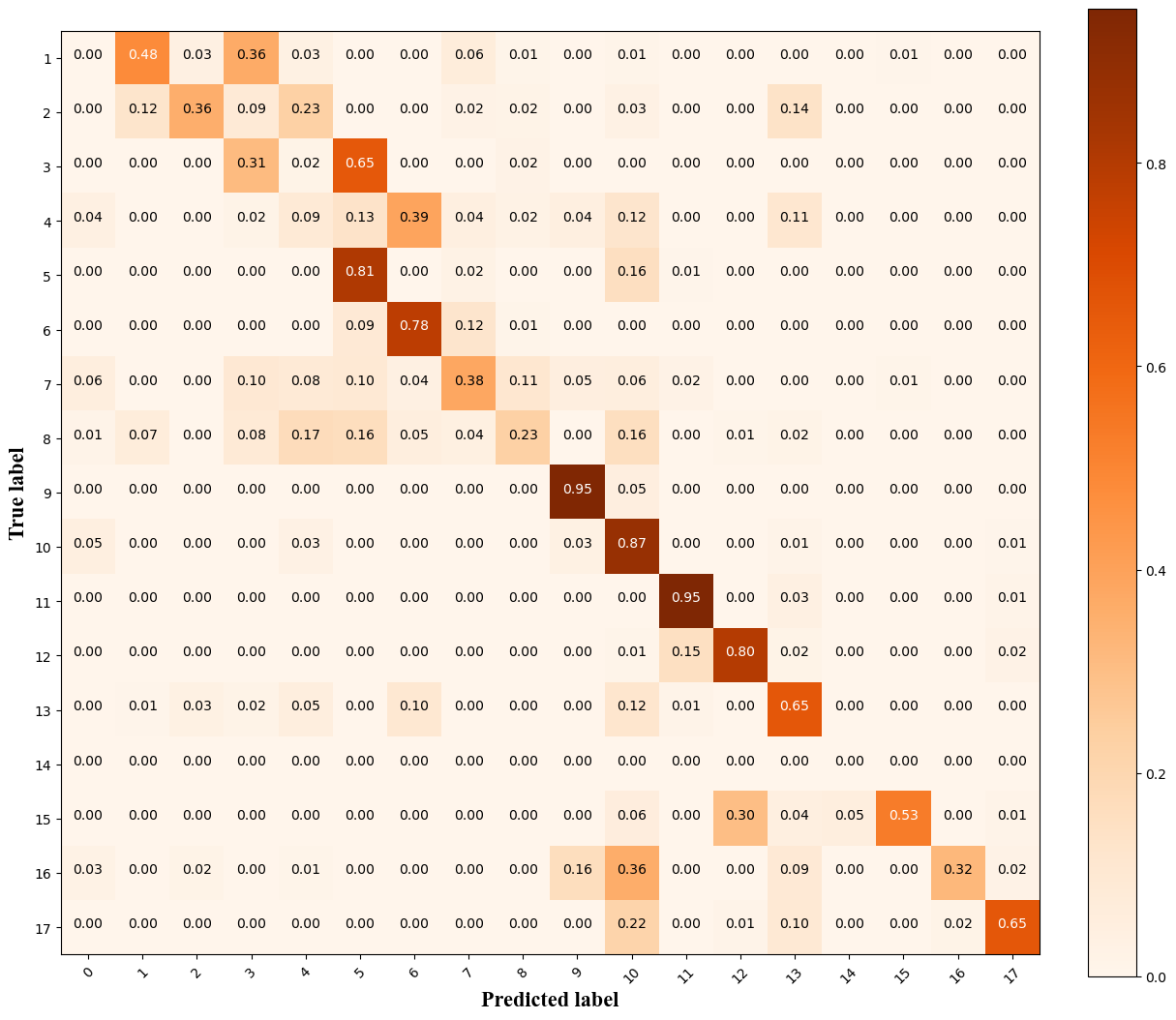}
    \caption{House A}
    \label{fig:cm_A}
    \end{subfigure}
\hfill
    \begin{subfigure}[b]{0.49\linewidth}
    \includegraphics[width=\linewidth]{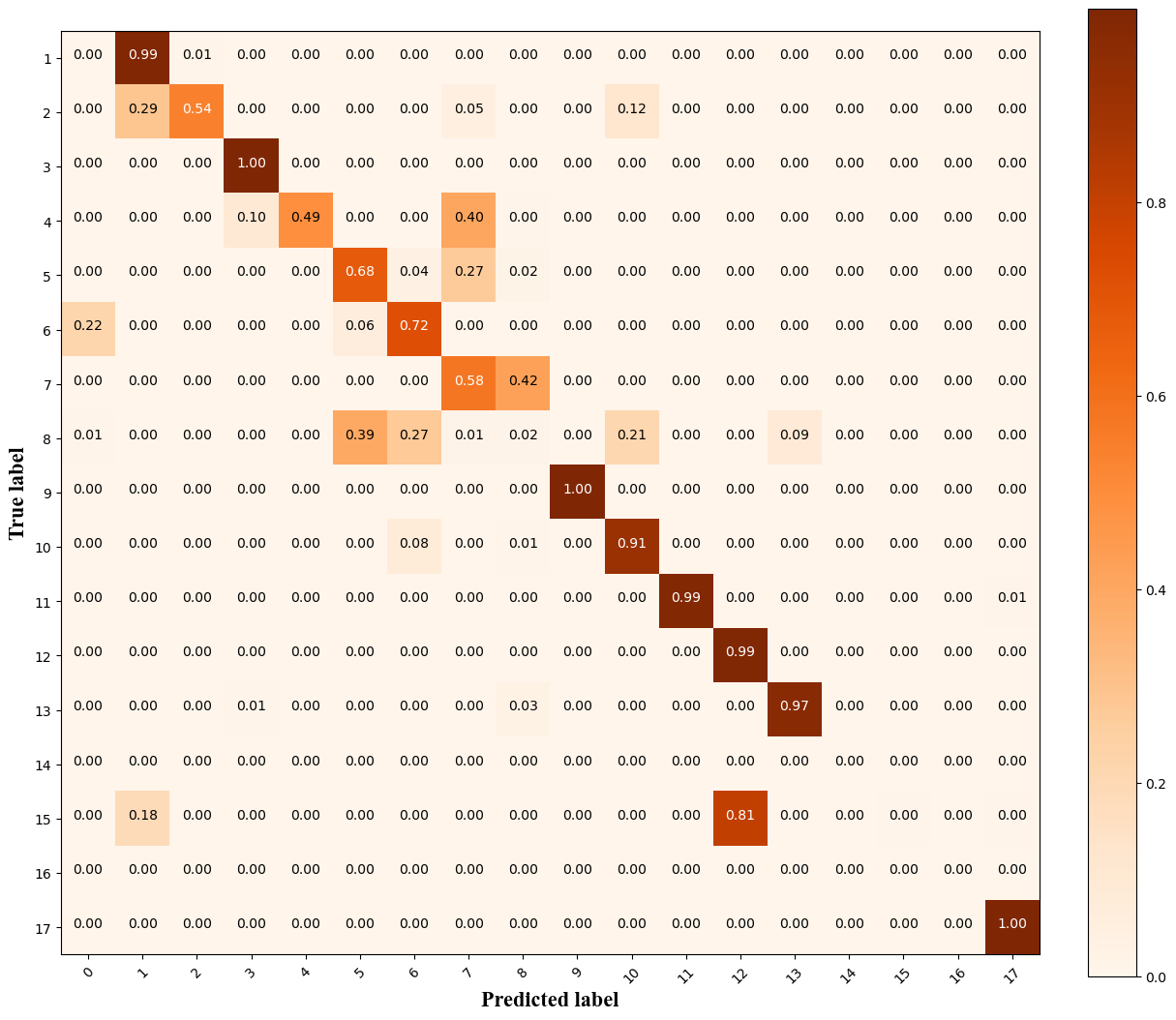}
    \caption{House B}
    \label{fig:cm_B}
    \end{subfigure}
\caption{Confusion matrices of single-user activity recognition.}
\label{fig:cm_A_B}
\end{figure}
 
\paragraph{Parameters}
Queries to the large language model are based on API calls to the gpt-4-32k-0613 model provided by the Azure OpenAI Service, with the Temperature parameter set to 0 to reduce the randomness of the model's output, while keeping other parameters at their default settings. For the two hyperparameters in LAHAR—the chunk size $N$ for the Action-Level Resident Separation and Description Generation module, and the chunk size $M$ for the Activity-Level Reasoning module—we used $N=20$ and $M=15$, respectively. This setup was determined based on preliminary experiments, taking into account two factors: on one hand, we need each chunk to contain as much context as possible, and on the other hand, chunks that are too long can impair the LLM's ability to follow instructions within the prompt.

\subsection{Evaluation Metric}

For a data segment whose time period is $T$, the activities occurred is denoted as $A_T = \{a_{k}^{T_j}\}_{j\in \mathcal{J}}$, where $a_k$ is $k$-th activity class in $K$ classes and $\mathcal{J} = \{j \in \mathbb{N} | T_j \subseteq T\}$. 
We perform one-hot encoding for all the activities $\{a_k\}$ present at each second of $T$ and apply the union operation. For example, if the $i$-th and $j$-th activities are ongoing at the second $t$, the encoding is a length-$K$ vector with ones at positions $i$ and $j$ and zeros elsewhere. By stacking all the seconds, we have a two-dimensional matrix $M_{T\times K}$. To compare our prediction $\hat{M}_{T\times K}$ with the ground truth $M_{T\times K}$, we have
\begin{align*}
    S_{T\times K} &= \hat{M}_{T\times K} \cdot M_{T\times K},\\
    TP &= \left[\sum_{t\in T}{S_{t,k}}\right]_{1\times K},\\
    FP &= \left[\sum_{t\in T}{(\hat{M}_{t,k}- S_{t,k})}\right]_{1\times K},\\
    FN &= \left[\sum_{t\in T}{(M_{t,k}- S_{t,k})}\right]_{1\times K}.\\
\end{align*}
Based on $TP$, $FP$, and $FN$ given above, we can then calculate the precision, recall, and F1-score of each class in our prediction.
\begin{table*}[t]
\centering
\caption{Results of activity recognition of each class in different scenarios}
\label{tab: f1_score}
 \resizebox{\textwidth}{!}{
\begin{tabular}{|cc|cccc|cccc|cccc|}
\hline
\multicolumn{2}{|c|}{Metric}                                         & \multicolumn{4}{c|}{Precision (\%)}                                                                        & \multicolumn{4}{c|}{Recall (\%)}                                                                           & \multicolumn{4}{c|}{F1-score (\%)}                                                                         \\ \hline
\multicolumn{2}{|c|}{Scenario}                                       & \multicolumn{1}{c|}{Single\_A} & \multicolumn{1}{c|}{Multi\_A} & \multicolumn{1}{c|}{Single\_B} & Multi\_B & \multicolumn{1}{c|}{Single\_A} & \multicolumn{1}{c|}{Multi\_A} & \multicolumn{1}{c|}{Single\_B} & Multi\_B & \multicolumn{1}{c|}{Single\_A} & \multicolumn{1}{c|}{Multi\_A} & \multicolumn{1}{c|}{Single\_B} & Multi\_B \\ \hline
\multicolumn{1}{|c|}{\multirow{15}{*}{Class}} & Preparing Breakfast  & \multicolumn{1}{c|}{59.95}     & \multicolumn{1}{c|}{66.54}    & \multicolumn{1}{c|}{67.41}     & 80.07    & \multicolumn{1}{c|}{50.45}     & \multicolumn{1}{c|}{62.17}    & \multicolumn{1}{c|}{98.12}     & 84.38    & \multicolumn{1}{c|}{54.79}     & \multicolumn{1}{c|}{64.28}    & \multicolumn{1}{c|}{79.92}     & 82.16    \\
\multicolumn{1}{|c|}{}                        & Having Breakfast     & \multicolumn{1}{c|}{65.05}     & \multicolumn{1}{c|}{97.70}    & \multicolumn{1}{c|}{97.29}     & 96.20    & \multicolumn{1}{c|}{39.21}     & \multicolumn{1}{c|}{74.18}    & \multicolumn{1}{c|}{53.93}     & 82.05    & \multicolumn{1}{c|}{48.92}     & \multicolumn{1}{c|}{84.33}    & \multicolumn{1}{c|}{69.40}     & 88.56    \\
\multicolumn{1}{|c|}{}                        & Preparing Lunch      & \multicolumn{1}{c|}{25.08}     & \multicolumn{1}{c|}{27.37}    & \multicolumn{1}{c|}{67.31}     & 73.46    & \multicolumn{1}{c|}{30.02}     & \multicolumn{1}{c|}{17.84}    & \multicolumn{1}{c|}{100.}       & 68.74    & \multicolumn{1}{c|}{27.33}     & \multicolumn{1}{c|}{21.60}    & \multicolumn{1}{c|}{80.46}     & 71.02    \\
\multicolumn{1}{|c|}{}                        & Having Lunch         & \multicolumn{1}{c|}{5.19}      & \multicolumn{1}{c|}{47.87}    & \multicolumn{1}{c|}{100.}       & 47.78    & \multicolumn{1}{c|}{9.00}      & \multicolumn{1}{c|}{33.33}    & \multicolumn{1}{c|}{62.22}     & 66.23    & \multicolumn{1}{c|}{6.58}      & \multicolumn{1}{c|}{39.30}    & \multicolumn{1}{c|}{76.71}     & 55.51    \\
\multicolumn{1}{|c|}{}                        & Preparing Dinner     & \multicolumn{1}{c|}{21.53}     & \multicolumn{1}{c|}{64.90}    & \multicolumn{1}{c|}{66.22}     & 54.56    & \multicolumn{1}{c|}{73.84}     & \multicolumn{1}{c|}{84.78}    & \multicolumn{1}{c|}{67.59}     & 76.56    & \multicolumn{1}{c|}{33.33}     & \multicolumn{1}{c|}{73.52}    & \multicolumn{1}{c|}{66.90}     & 63.71    \\
\multicolumn{1}{|c|}{}                        & Having Dinner        & \multicolumn{1}{c|}{7.59}      & \multicolumn{1}{c|}{58.92}    & \multicolumn{1}{c|}{11.67}     & 31.39    & \multicolumn{1}{c|}{82.47}     & \multicolumn{1}{c|}{65.67}    & \multicolumn{1}{c|}{70.04}     & 29.91    & \multicolumn{1}{c|}{13.89}     & \multicolumn{1}{c|}{62.11}    & \multicolumn{1}{c|}{20.01}     & 30.63    \\
\multicolumn{1}{|c|}{}                        & Washing Dishes       & \multicolumn{1}{c|}{46.22}     & \multicolumn{1}{c|}{41.06}    & \multicolumn{1}{c|}{3.85}      & 10.37    & \multicolumn{1}{c|}{33.02}     & \multicolumn{1}{c|}{12.57}    & \multicolumn{1}{c|}{57.55}     & 7.97     & \multicolumn{1}{c|}{38.52}     & \multicolumn{1}{c|}{19.25}    & \multicolumn{1}{c|}{7.21}      & 9.00     \\
\multicolumn{1}{|c|}{}                        & Having Snack         & \multicolumn{1}{c|}{40.53}     & \multicolumn{1}{c|}{48.07}    & \multicolumn{1}{c|}{0.58}      & 12.84    & \multicolumn{1}{c|}{22.04}     & \multicolumn{1}{c|}{26.27}    & \multicolumn{1}{c|}{1.54}      & 44.42    & \multicolumn{1}{c|}{28.56}     & \multicolumn{1}{c|}{33.97}    & \multicolumn{1}{c|}{0.85}      & 19.92    \\
\multicolumn{1}{|c|}{}                        & Sleeping             & \multicolumn{1}{c|}{87.84}     & \multicolumn{1}{c|}{95.53}    & \multicolumn{1}{c|}{93.80}     & 97.84    & \multicolumn{1}{c|}{93.99}     & \multicolumn{1}{c|}{88.20}    & \multicolumn{1}{c|}{100}       & 97.82    & \multicolumn{1}{c|}{90.81}     & \multicolumn{1}{c|}{91.72}    & \multicolumn{1}{c|}{96.80}     & 97.83    \\
\multicolumn{1}{|c|}{}                        & Entertainment        & \multicolumn{1}{c|}{70.44}     & \multicolumn{1}{c|}{69.31}    & \multicolumn{1}{c|}{98.58}     & 89.45    & \multicolumn{1}{c|}{88.65}     & \multicolumn{1}{c|}{90.66}    & \multicolumn{1}{c|}{90.85}     & 90.28    & \multicolumn{1}{c|}{78.50}     & \multicolumn{1}{c|}{78.56}    & \multicolumn{1}{c|}{94.55}     & 89.86    \\
\multicolumn{1}{|c|}{}                        & Having Shower        & \multicolumn{1}{c|}{72.95}     & \multicolumn{1}{c|}{59.39}    & \multicolumn{1}{c|}{100}       & 60.27    & \multicolumn{1}{c|}{91.82}     & \multicolumn{1}{c|}{81.05}    & \multicolumn{1}{c|}{84.99}     & 86.57    & \multicolumn{1}{c|}{81.30}     & \multicolumn{1}{c|}{68.55}    & \multicolumn{1}{c|}{91.89}     & 71.06    \\
\multicolumn{1}{|c|}{}                        & Toileting            & \multicolumn{1}{c|}{82.67}     & \multicolumn{1}{c|}{53.74}    & \multicolumn{1}{c|}{82.40}     & 91.37    & \multicolumn{1}{c|}{76.86}     & \multicolumn{1}{c|}{81.43}    & \multicolumn{1}{c|}{96.76}     & 84.27    & \multicolumn{1}{c|}{79.66}     & \multicolumn{1}{c|}{64.75}    & \multicolumn{1}{c|}{89.00}     & 87.68    \\
\multicolumn{1}{|c|}{}                        & Working              & \multicolumn{1}{c|}{76.60}     & \multicolumn{1}{c|}{81.50}    & \multicolumn{1}{c|}{99.51}     & 98.39    & \multicolumn{1}{c|}{61.08}     & \multicolumn{1}{c|}{60.49}    & \multicolumn{1}{c|}{98.81}     & 85.63    & \multicolumn{1}{c|}{67.96}     & \multicolumn{1}{c|}{69.44}    & \multicolumn{1}{c|}{99.16}     & 91.57    \\
\multicolumn{1}{|c|}{}                        & Shaving              & \multicolumn{1}{c|}{/}         & \multicolumn{1}{c|}{84.16}    & \multicolumn{1}{c|}{/}         & 0.        & \multicolumn{1}{c|}{/}         & \multicolumn{1}{c|}{53.17}    & \multicolumn{1}{c|}{/}         & 0.       & \multicolumn{1}{c|}{/}         & \multicolumn{1}{c|}{65.17}    & \multicolumn{1}{c|}{/}         & 0.        \\
\multicolumn{1}{|c|}{}                        & Brushing Teeth       & \multicolumn{1}{c|}{75.24}     & \multicolumn{1}{c|}{46.63}    & \multicolumn{1}{c|}{1.09}      & 84.32    & \multicolumn{1}{c|}{43.86}     & \multicolumn{1}{c|}{34.30}    & \multicolumn{1}{c|}{0.40}      & 51.15    & \multicolumn{1}{c|}{55.42}     & \multicolumn{1}{c|}{39.53}    & \multicolumn{1}{c|}{0.58}      & 63.68    \\
\multicolumn{1}{|l|}{}                        & Talking on the Phone & \multicolumn{1}{c|}{98.67}     & \multicolumn{1}{c|}{42.83}    & \multicolumn{1}{c|}{/}         & 0.        & \multicolumn{1}{c|}{29.56}     & \multicolumn{1}{c|}{33.96}    & \multicolumn{1}{c|}{/}         & 0.        & \multicolumn{1}{c|}{45.49}     & \multicolumn{1}{c|}{37.88}    & \multicolumn{1}{c|}{/}         & 0.        \\
\multicolumn{1}{|l|}{}                        & Changing Clothes     & \multicolumn{1}{c|}{49.93}     & \multicolumn{1}{c|}{64.54}    & \multicolumn{1}{c|}{83.07}     & 95.19    & \multicolumn{1}{c|}{52.56}     & \multicolumn{1}{c|}{59.45}    & \multicolumn{1}{c|}{92.06}     & 82.95    & \multicolumn{1}{c|}{51.22}     & \multicolumn{1}{c|}{61.89}    & \multicolumn{1}{c|}{87.34}     & 88.65    \\ \hline
\multicolumn{2}{|c|}{Macro-Average}                                  & \multicolumn{1}{c|}{55.34}     & \multicolumn{1}{c|}{61.77}    & \multicolumn{1}{c|}{64.85}          & 60.20    & \multicolumn{1}{c|}{54.90}     & \multicolumn{1}{c|}{56.44}    & \multicolumn{1}{c|}{71.66}          & 61.11    & \multicolumn{1}{c|}{50.14}     & \multicolumn{1}{c|}{57.40}    & \multicolumn{1}{c|}{64.05}          & 59.46    \\ \hline
\multicolumn{2}{|c|}{Weighted-Average}                                  & \multicolumn{1}{c|}{66.00}     & \multicolumn{1}{c|}{77.67}    & \multicolumn{1}{c|}{92.37}     & 93.60    & \multicolumn{1}{c|}{67.56}     & \multicolumn{1}{c|}{76.23}    & \multicolumn{1}{c|}{93.79}     & 90.28    & \multicolumn{1}{c|}{76.95}     & \multicolumn{1}{c|}{86.49}    & \multicolumn{1}{c|}{93.08}     & 91.91    \\ \hline
\end{tabular}}
\end{table*}

\subsection{Single-Subject Activity Recognition}

To validate the activity recognition capability of LAHAR, a comparison is performed against the research of Gao et al. \cite{gao2024unsupervised}.  The experimental setup for this comparison is consistent with the research of Gao el al., focusing solely on the recognition of selected nine activity categories in single-person scenarios. We extract the longest data segments from the original data where Resident 2 was leaving home and Resident 1's activities are in these nine activities. These segments vary in length and can contain more than one activity. Without additional segmentation, our method can generate action-level descriptions and achieve activity sequence prediction at the resolution of a second. In contrast, the method of Gao et al, which did not use events as the smallest divisible units but instead used a fixed 5-minute time window, has thus a resolution of 5 minutes. Despite our higher resolution, our results are comparable to the results of Gao et al. in terms of the confusion matrix in Figure \ref{fig:cm_compare} and of precision, recall, and F1 score as shown in Table \ref{tab:comparison}. 

\paragraph{Extended Validation}

We further validated our method in more realistic and complex scenarios. As described in the experiment settings, our single-person scenario data include 17 activities across both houses, without class-based segmenting. Consequently, each segment is longer and contains more actions, making it more complex compared to the setting of Gao et al. \cite{gao2024unsupervised}.

Figure \ref{fig:cm_A_B} presents the confusion matrices of both houses. Furthermore, Table \ref{tab: f1_score} gives the precision, recall, and F1 scores of each class. 
From the confusion matrix of House A, it can be observed that compared to extracting data segments for 9 activities individually in Gao's setting, having more categories and longer data segments results in a lag in predicting activities related to breakfast and lunch. One main reason is the issue of activity alternation. For example, if a subject briefly watches TV while preparing breakfast and then resumes breakfast preparation, the LLM might interpret this as the subject having already prepared breakfast earlier and now preparing lunch. 
In House B, the activity of brushing teeth is difficult to accurately recognize because the subjects habitually use the toilet after brushing their teeth. This leads the LLM to merge and predict brushing teeth and using the toilet as a single activity of toileting.

\subsection{Multi-Subject Activity Recogntion}
Since the ARAS dataset does not label events with the IDs of their subjects, we cannot perform a one-to-one comparison of event assignments. To validate our method's activity recognition capability in multi-person scenarios, we qualitatively present an example of our results and indirectly demonstrate our method's ability to separate residents by comparing its performance with that in single-person scenarios.

\paragraph{Qualitative Results}

Figure \ref{fig:qulitative_example} provides an example to better illustrate our results. We excerpted the experimental output of about 21 minutes of sensor data from 22:15:24 to 22:36:41 of the 5th day in House B in the multi-person scenario. It can be seen that these sensor events were integrated into 9 descriptions by the LLM, in which each description is assigned to a subject. By separating these descriptions by subjects, timestamped activities are finally predicted respectively for each resident.

\paragraph{Quantitative Results}
In Table \ref{tab: f1_score}, the recognition results for each activity class in multi-person scenarios are presented. It can be seen that even when extended to multiple people, our method's performance in activity recognition remains comparable to that in single-person scenarios. Although the time and activity distributions differ between single-person and multi-person scenarios, this comparison still highlights the scalability of LAHAR in multi-person contexts.

\section{Conclusion}
In this paper, we propose LAHAR, a framework using LLMs for multi-person HAR with ambient sensors. Our prompts enable LLMs to assign sensor events to individuals based on their states, generating detailed descriptions and reasoning about their activities. This method extends LLM application to multi-person HAR, achieving time resolutions matching sensor timestamps. LAHAR's explicit descriptions and activity reasoning offer promising perspectives to address explainability challenges. Experimental validation shows performance comparable to the state-of-the-art in single-person and multi-person scenarios. Future plans include validation with different LLMs, model fine-tuning, and further evaluation of conversational explainability.

\bibliographystyle{splncs04}
\bibliography{biblio}

\end{document}